\documentclass{article}

\PassOptionsToPackage{numbers, sort&compress}{natbib}

\usepackage[preprint]{neurips_2023}

\usepackage{natbib}
\usepackage{wrapfig}
\usepackage{times}
\usepackage{latexsym}
\usepackage{tablefootnote}
\usepackage{url}
\usepackage[T1]{fontenc}
\usepackage[utf8]{inputenc}
\usepackage{microtype}
\usepackage{graphicx}
\usepackage{subfigure}

\usepackage{algorithm}
\usepackage{listings}
\usepackage{inconsolata}
\usepackage{times}
\usepackage{latexsym}
\usepackage{bm}
\usepackage{amsthm,amsmath,amssymb}
\usepackage{mathrsfs}
\usepackage{epsfig}
\usepackage{booktabs}
\usepackage{verbatim}
\usepackage{enumitem,kantlipsum}
\usepackage{lipsum}
\usepackage{tipa}
\usepackage{algorithmicx}
\usepackage{algpseudocode}
\usepackage{multirow}
\usepackage{arydshln}
\usepackage{xcolor}
\usepackage{caption}
\usepackage{hyperref}

\usepackage{color} 
\usepackage{dsfont}
\usepackage{tikz}

\usepackage{cleveref}
\crefname{section}{§}{§§}
\Crefname{section}{§}{§§}

\definecolor{NavyBlue}{rgb}{0.1, 0.4, 0.8}
\hypersetup{colorlinks = true, linkcolor = NavyBlue,
            urlcolor  = gray,
            citecolor = NavyBlue,
            anchorcolor = NavyBlue}

\title{PandaGPT: \\ One Model To Instruction-Follow Them All}

\author{%
  Yixuan Su$^{\spadesuit,}$\thanks{Major contributors. Contact: \texttt{ys484@cam.ac.uk} and \texttt{jcykcai@tencent.com}.}\ \ $^{,}$\thanks{Work done during internship at Tencent AI Lab.}  \And Tian Lan$^*$ \And Huayang Li$^{\diamondsuit,}$\footnotemark[1]\ \ $^{,}$\footnotemark[2] \And Jialu Xu \And Yan Wang \And Deng Cai${^{\clubsuit,}}$$^*$ \\
  \\
   $^\spadesuit$University of Cambridge\ \ \ \ \ $^\diamondsuit$Nara Institute of Science and Technology\\
   $^\clubsuit$Tencent AI Lab\\
  \AND
  \url{https://panda-gpt.github.io/} \\
}

\begin{document}

\maketitle

\begin{abstract}
We present PandaGPT, an approach to em\underline{\textbf{P}}ower large l\underline{\textbf{AN}}guage mo\underline{\textbf{D}}els with visual and \underline{\textbf{A}}uditory instruction-following capabilities. 
Our pilot experiments show that PandaGPT can perform complex tasks such as detailed image description generation, writing stories inspired by videos, and answering questions about audios. More interestingly, PandaGPT can take multimodal inputs simultaneously and compose their semantics naturally. For example, PandaGPT can connect how objects look in an image/video and how they sound in an audio. To do so, PandaGPT combines the multimodal encoders from ImageBind and the large language models from Vicuna. Notably, only aligned image-text pairs are required for the training of PandaGPT. Thanks to the strong capability of ImageBind in embedding data from different modalities into the same space, PandaGPT displays emergent, i.e. zero-shot, cross-modal behaviors for data other than image and text (e.g., video, audio, depth, thermal, and IMU). 
We hope that PandaGPT serves as an initial step toward building AGI that can perceive and understand inputs in different modalities holistically, as we humans do.
\end{abstract}

\section{Introduction}
Humans possess remarkable abilities to perceive and understand information from diverse sensory modalities, such as seeing a painting and hearing an audio guide. Analogously, to learn simultaneously, holistically, and directly from many different forms of information holds great promise for enabling machines to have a more comprehensive and better understanding of the world. To this end, there has been an emergent interest in developing artificial intelligence (AI) systems capable of perceiving and understanding information from multiple modalities simultaneously in a manner similar to humans.

However, much of the prior research has focused on tackling individual modalities in isolation. For instance, while significant progress has been made in text-to-image retrieval and generation \cite{radford2021learning}, visually-grounded instruction following \cite{liu2023llava,zhu2023minigpt}, and speech understanding and generation \cite{zhang2023speechgpt}, these advances have largely been confined to separate combinations of text and other modalities or, at best, a few visual modalities (e.g., image and video). These models are limited in their ability to connect information from different modalities and lack the capacity to perceive and understand multimodal inputs holistically, thereby neglecting the inherent richness and complementary nature of multimodal data.

In this paper, we present PandaGPT, the first general-purpose model capable of instruction-following data from six modalities. 
PandaGPT leverages the power of multimodal encoders from ImageBind \cite{girdhar2023imagebind} and the expressive language models from Vicuna \cite{vicuna2023}, demonstrating impressive and emergent cross-modal capabilities across six modalities: image/video, text, audio, depth, thermal, and inertial measurement units (IMU). Crucially, PandaGPT achieves these capabilities despite being only trained on aligned image-text pairs, thanks to the shared embedding space provided by ImageBind.

This integration of multimodal information enables PandaGPT to perform a wide range of tasks, including generating detailed descriptions of images, composing engaging stories inspired by videos, and providing accurate answers to questions about audio inputs. Most interestingly, the core innovation of PandaGPT lies in its ability to naturally compose the semantics of multimodal inputs, which enables a rich set of compositional multimodal tasks across different modalities. For example, it can seamlessly connect the visual appearance of objects in a photo with their corresponding sounds in an audio clip, producing a cohesive and comprehensive understanding of the scene. These cross-modal capabilities empower the model to go beyond traditional unimodal analysis. We hope PandaGPT serves as an initial step toward building AGI that can perceive and understand inputs in different modalities holistically, as humans do.

\section{Related Work}
\paragraph{Large Language Models.} Large language models (LLMs) pre-trained over massive unlabeled text have dominated the field of natural language processing (NLP) today \cite{radford2018improving,devlin2019bert,radford2019language,brown2020language}. With alignment techniques such as supervised instruction tuning \cite{sanh2021multitask,wei2021finetuned,mishra2021natural} and reinforcement learning from human feedback \cite{stiennon2020learning,ouyang2022training}, LLMs exhibit surprisingly effective zero- and few-shot generalization abilities to perform almost any NLP tasks. The most successful examples could be OpenAI's ChatGPT \cite{chatgpt} and GPT4 \cite{gpt4}, which have made a profound impact on the entire AI research community and beyond. There also have been enormous open-source efforts to replicate the success, such as BLOOM \cite{scao2022bloom}, LLaMA \cite{touvron2023llama}, Alpaca \cite{alpaca}, Vicuna \cite{vicuna2023}, OpenAlpaca \cite{openalpaca} among many others.

\paragraph{Multi-modal Alignment.}
Feature alignment among multiple modalities has attracted great interest for its applications such as cross-modal retrieval \cite{frome2013devise,faghri2017vse++,alayrac2020self}. Recently, CLIP \cite{radford2021learning} learns a joint embedding space for image and text. Flamingo \cite{alayrac2022flamingo}, BLIP-2 \cite{li2023blip}, and MAGIC~\citep{su2022language} bridge powerful pre-trained vision-only and language-only models and show strong zero-shot abilities. AudioCLIP \cite{guzhov2022audioclip} adds audio into the CLIP framework for audio classification. ImageBind \cite{girdhar2023imagebind} learn a joint embedding across six different modalities (image/video, text, audio, depth, thermal, and IMU data) using image-paired data only. More recently, there has been a surge of interest to combine multi-modal alignment and large language models for multi-modal instruction following. 
LLaVa \cite{liu2023llava}, Mini-GPT4 \cite{zhu2023minigpt}, and Video-LLaMA~\citep{damonlpsg2023videollama} enable visually-grounded instruction following. DetGPT~\citep{detgpt2023} proposes reasoning-based object detection. SpeechGPT \cite{zhang2023speechgpt} adds speech understanding and generation abilities to LLMs. However, these advances have largely been confined to separate combinations of text and other modalities (e.g., image/video or audio).  

\begin{figure*}[h]
\centering
  \includegraphics[width=0.9\linewidth]{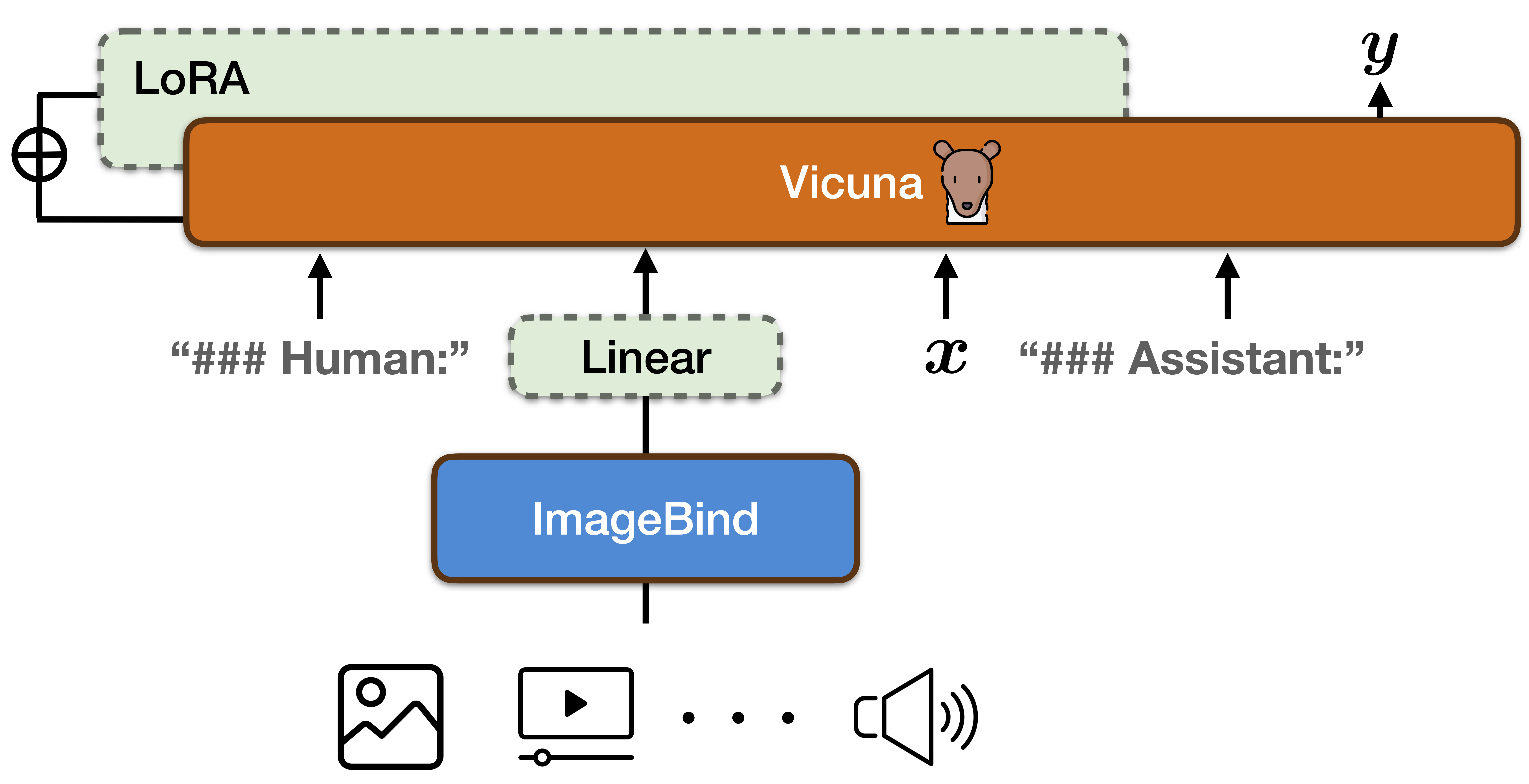}
  \caption{Illustration of PandaGPT. During training, we only train the linear projection matrix and the additional LoRA weights (as indicated with dashed boxes) while keeping the parameters of ImageBind and Vicuna frozen.}
  \label{fig:pandagpt}
\end{figure*}

\section{Method}
PandaGPT combines the multi-modal encoders from ImageBind and the large language models from Vicuna, achieving impressive capabilities in vision- and audio-grounded instruction following tasks. To align the feature space of multimodal encoders from ImageBind and large language models from Vicuna\footnote{We use the version-0 of Vicuna-13B as our base language model.}, we train PandaGPT using 160k image-language instruction-following data released by~\citep{liu2023llava} and~\citep{zhu2023minigpt}. Each training instance consists of an image $\mathcal{I}$ and a multi-turn conversation data $(\boldsymbol{x}_1, \boldsymbol{y}_1, ..., \boldsymbol{x}_n, \boldsymbol{y}_n)$, where $\boldsymbol{x}_i$ and $\boldsymbol{y}_i$ are the human's instruction and the system's response at the $i$-th turn. To reduce the number of trainable parameters, we only train (i) a linear projection matrix $f$ to connect the representation produced by ImageBind to Vicuna; and (ii) additional LoRA~\citep{hu2021lora} weights on the Vicuna's attention modules.\footnote{The total number of trainable parameters is around 0.4\% of the parameters of Vicuna.} Figure~\ref{fig:pandagpt} illustrates the architecture of PandaGPT. 

The training objective of PandaGPT is defined as 
\begin{equation}
    \mathcal{L}(\theta_{f}, \theta_{l}) = \prod_{i=1}^{n}p_{\theta}(\boldsymbol{y}_i|\boldsymbol{x}_{<i}, \boldsymbol{y}_{<i-1}, f(h_{\mathcal{I}})),
\end{equation}
where $\theta_{f}$ and $\theta_{l}$ correspond to the learnable parameters of the linear projection matrix and LoRA weights. The $h_{\mathcal{I}}$ is the image representation produced by ImageBind and $\theta=\{\theta_{f}, \theta_{l}, \theta_{1}, \theta_{2}\}$, where $\theta_{1}$ and $\theta_{2}$ are frozen parameters of ImageBind and Vicuna. Note that the loss is only computed from the part of system responses during training. We train PandaGPT on the image-language instruction-following dataset for two epochs using a learning rate of 5e-4 with linear decay. The maximum sequence length for Vicuna-13B is set to 400 based on our computation resources (8$\times$A100 40G GPUs). The training takes around 7 hours to complete.



It is worth noting that the current version of PandaGPT is only trained with aligned image-text data. However, by leveraging the binding property across six modalities (image/video, text, audio, depth, thermal, and IMU) inherited from the frozen ImageBind encoders, PandaGPT demonstrates emergent, i.e. zero-shot, cross-modal capabilities across all of the modalities.

\section{Capabilities of PandaGPT}
\begin{figure*}[h]
\centering
  \includegraphics[width=0.9\linewidth]{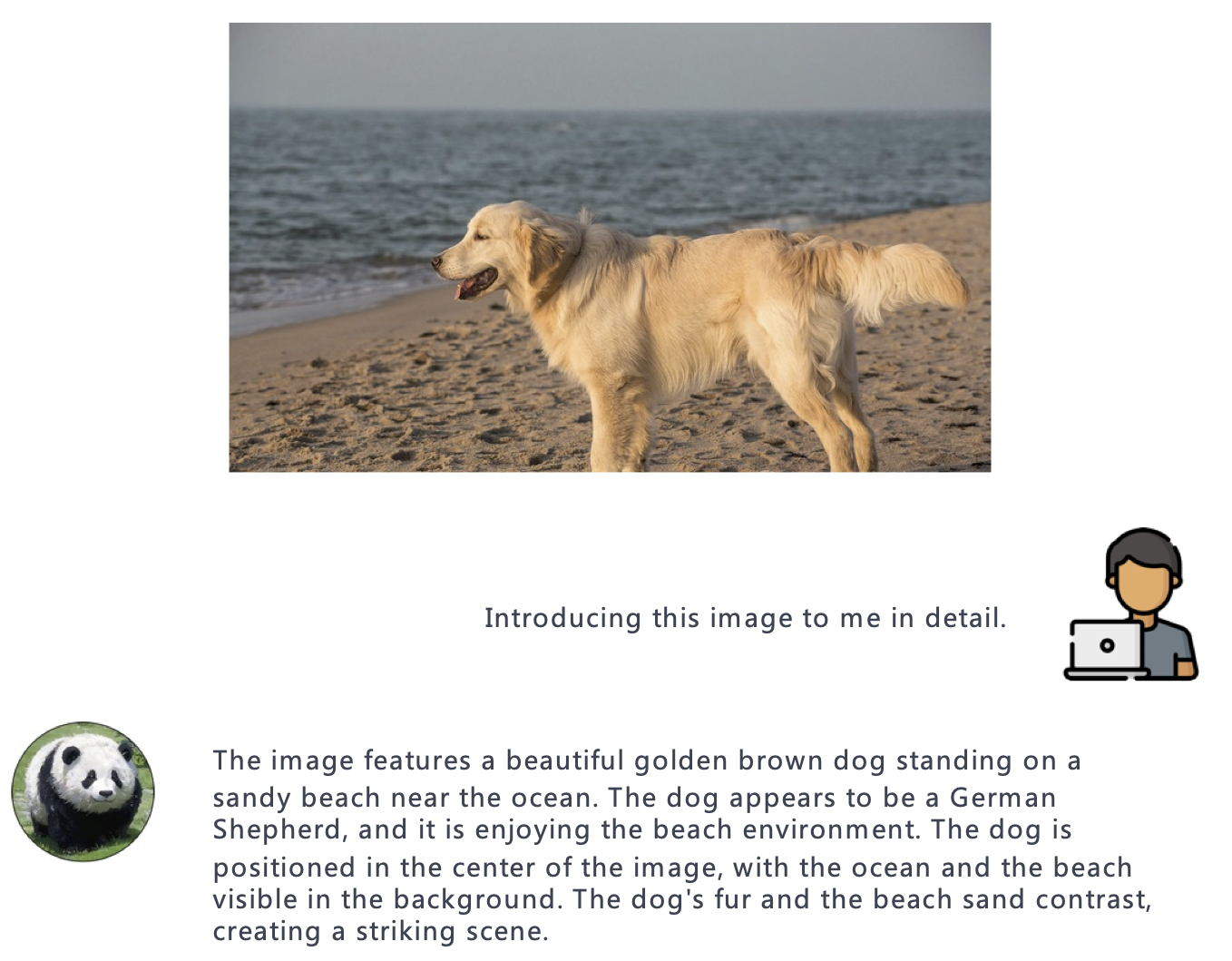}
  \caption{Example showing PandaGPT's capability in image-grounded question answering.}
  \label{fig:image-dog}
\end{figure*}

\begin{figure*}[h]
\centering
  \includegraphics[width=0.9\linewidth]{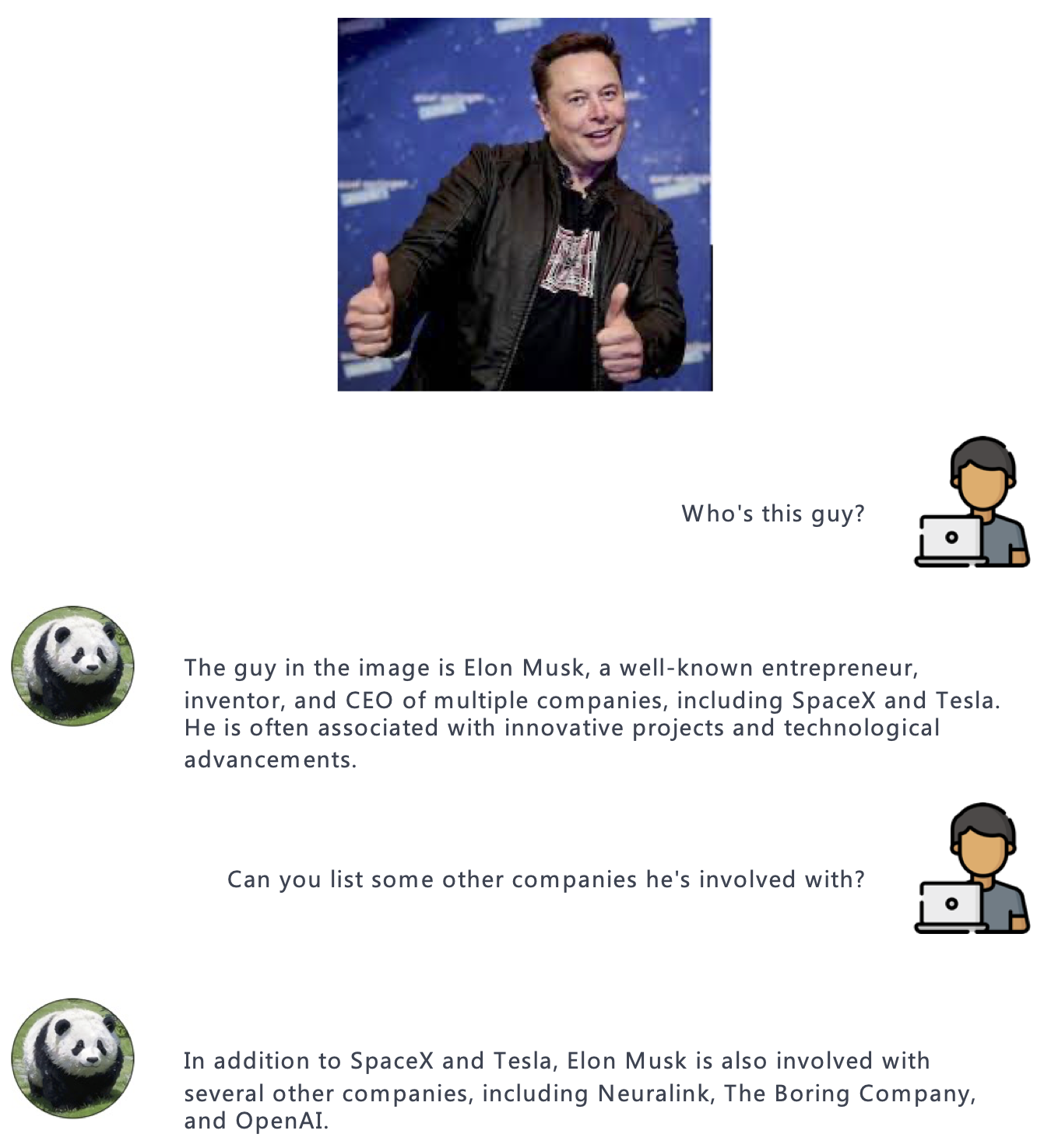}
  \caption{Example showing PandaGPT's capability in image-grounded question answering.}
  \label{fig:image-musk}
\end{figure*}

\begin{figure*}[h]
\centering
  \includegraphics[width=0.9\linewidth]{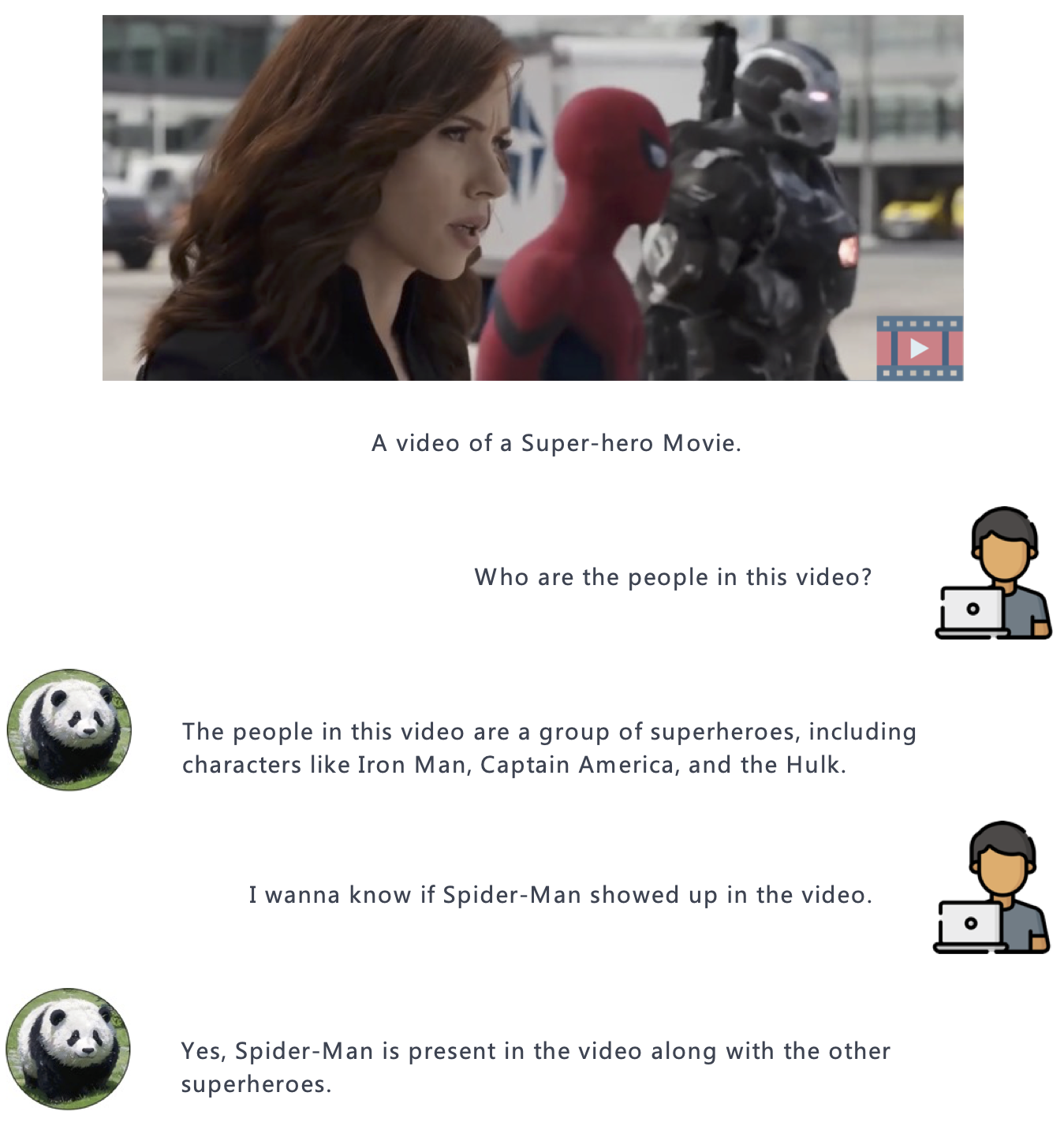}
  \caption{Example showing PandaGPT's capability in video-grounded question answering.}
  \label{fig:video-avengers}
\end{figure*}

\begin{figure*}[h]
\centering
  \includegraphics[width=0.9\linewidth]{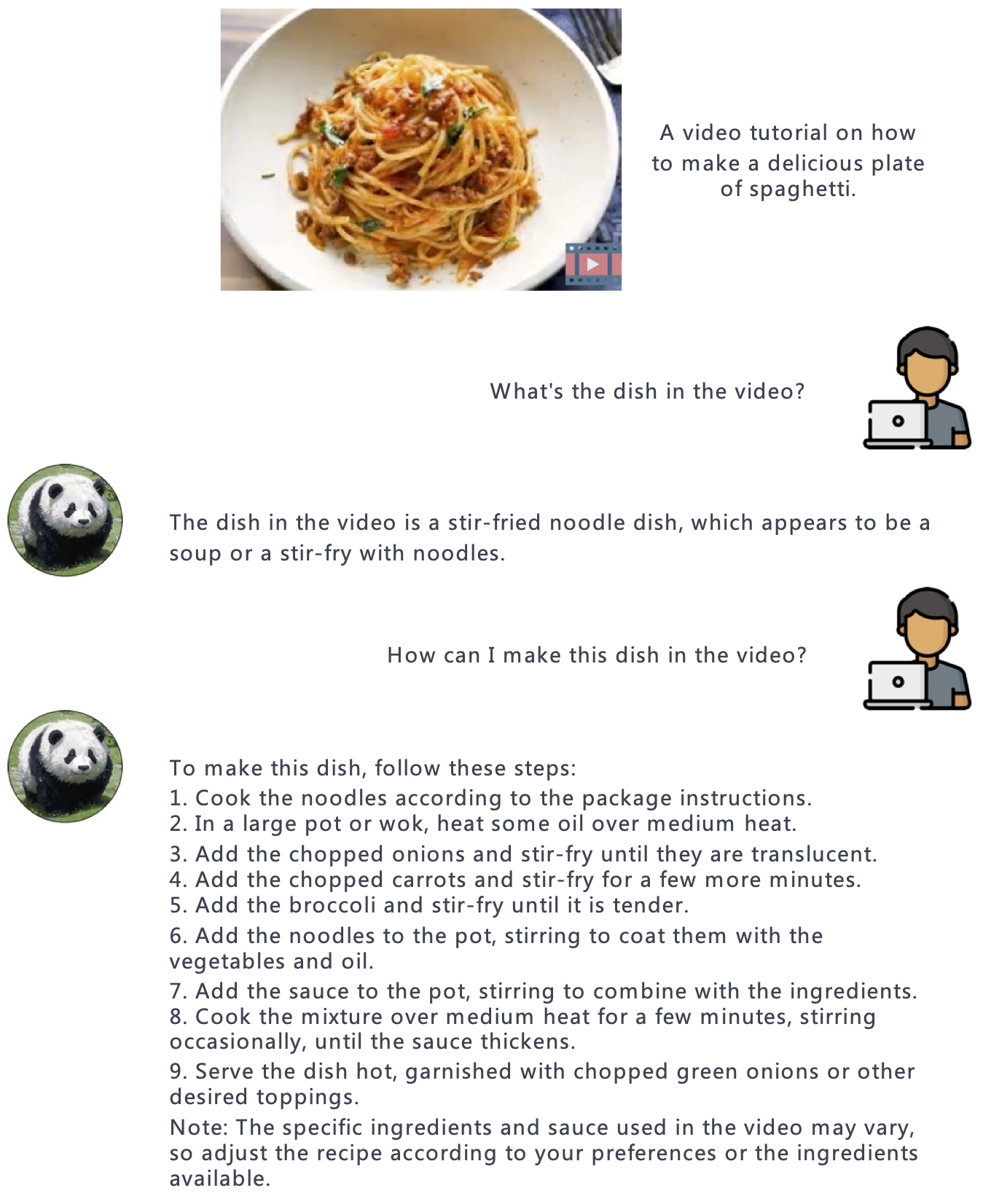}
  \caption{Example showing PandaGPT's capability in image/video-inspired creative writing.}
  \label{fig:video-noodle}
\end{figure*}

\begin{figure*}[h]
\centering
  \includegraphics[width=0.9\linewidth]{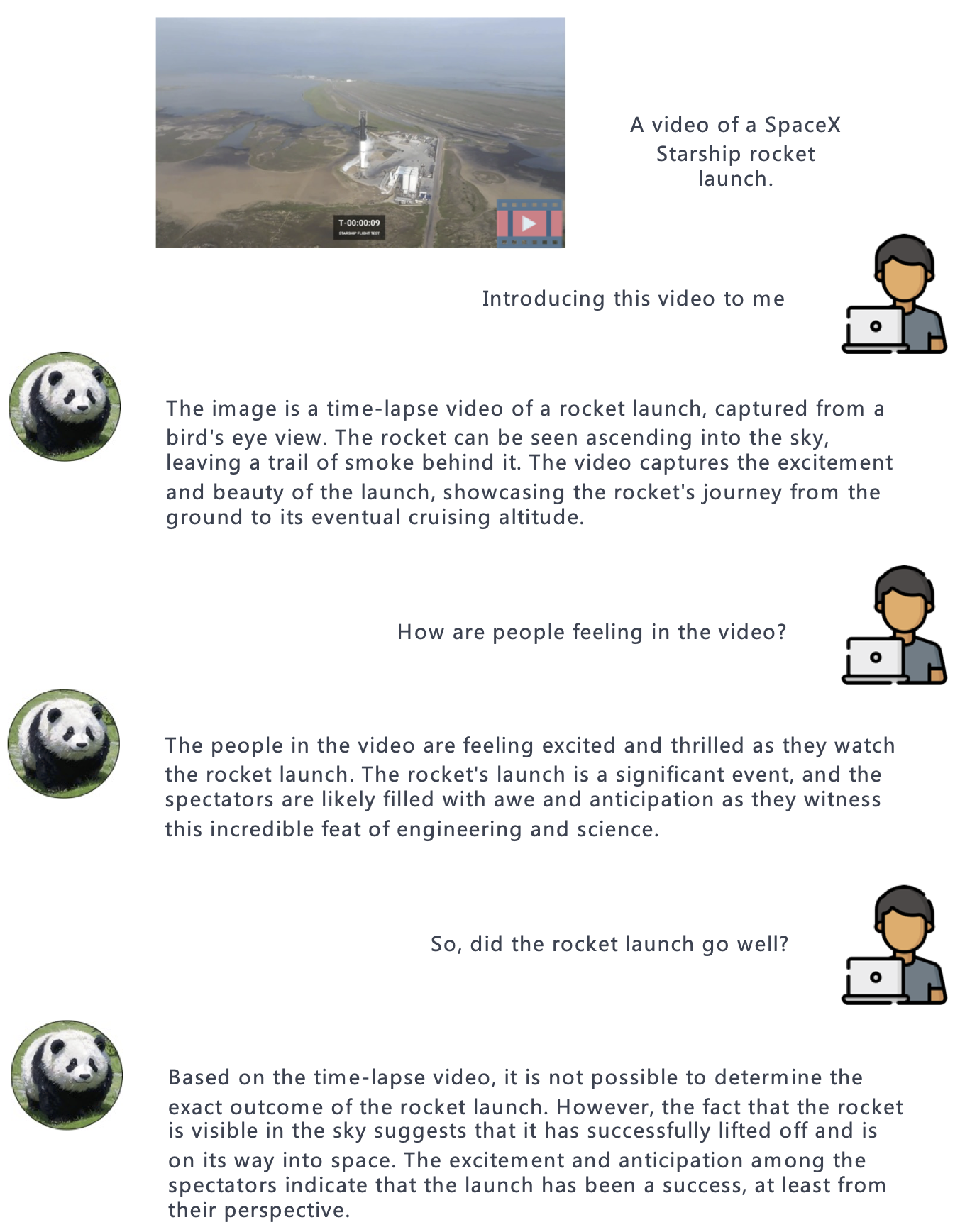}
  \caption{Example showing PandaGPT's capability in visual reasoning.}
  \label{fig:video-spacex-rocket}
\end{figure*}

\begin{figure*}[h]
\centering
  \includegraphics[width=0.9\linewidth]{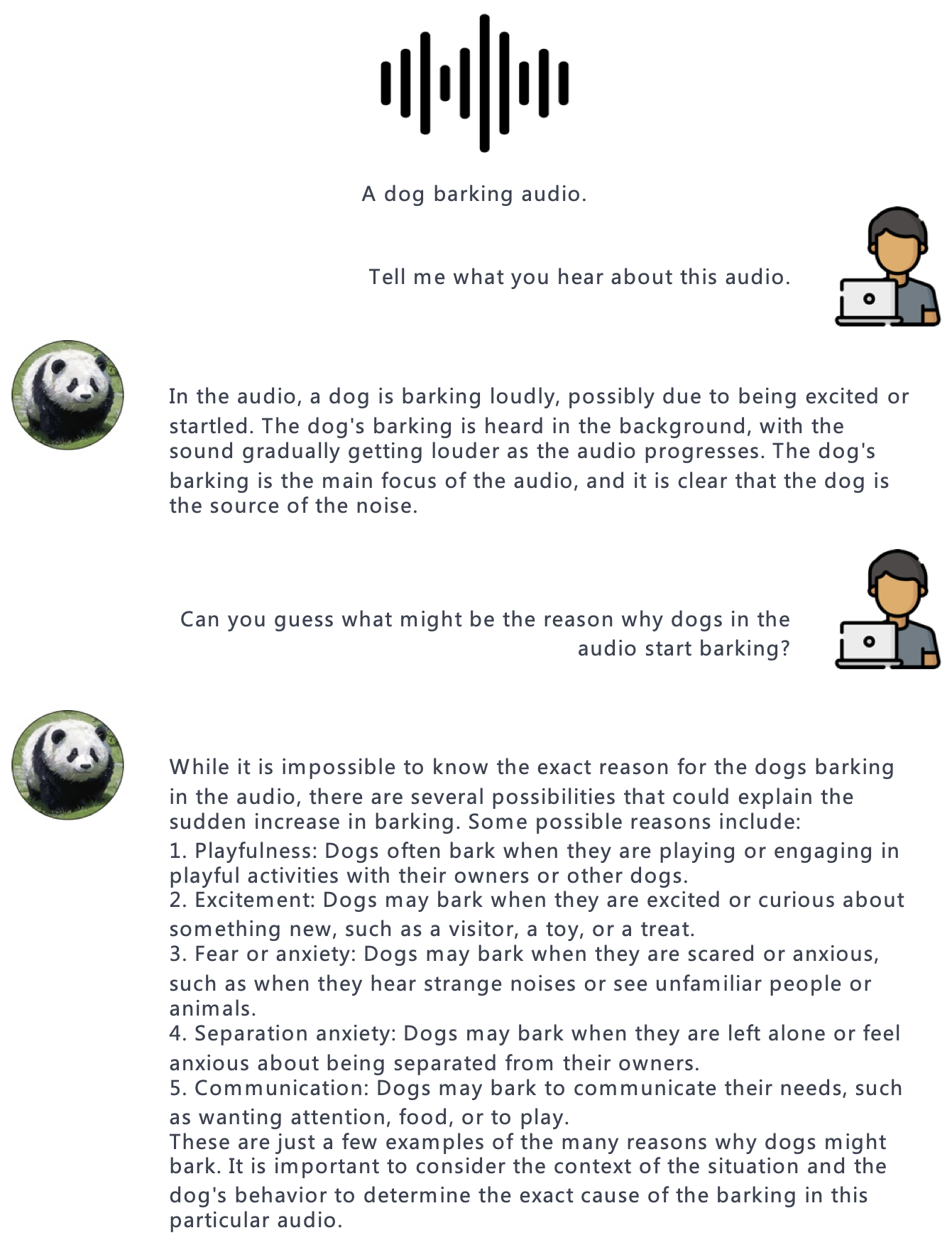}
  \caption{Example showing PandaGPT's capability in auditory reasoning.}
  \label{fig:audio-dog}
\end{figure*}

\begin{figure*}[h]
\centering
  \includegraphics[width=0.9\linewidth]{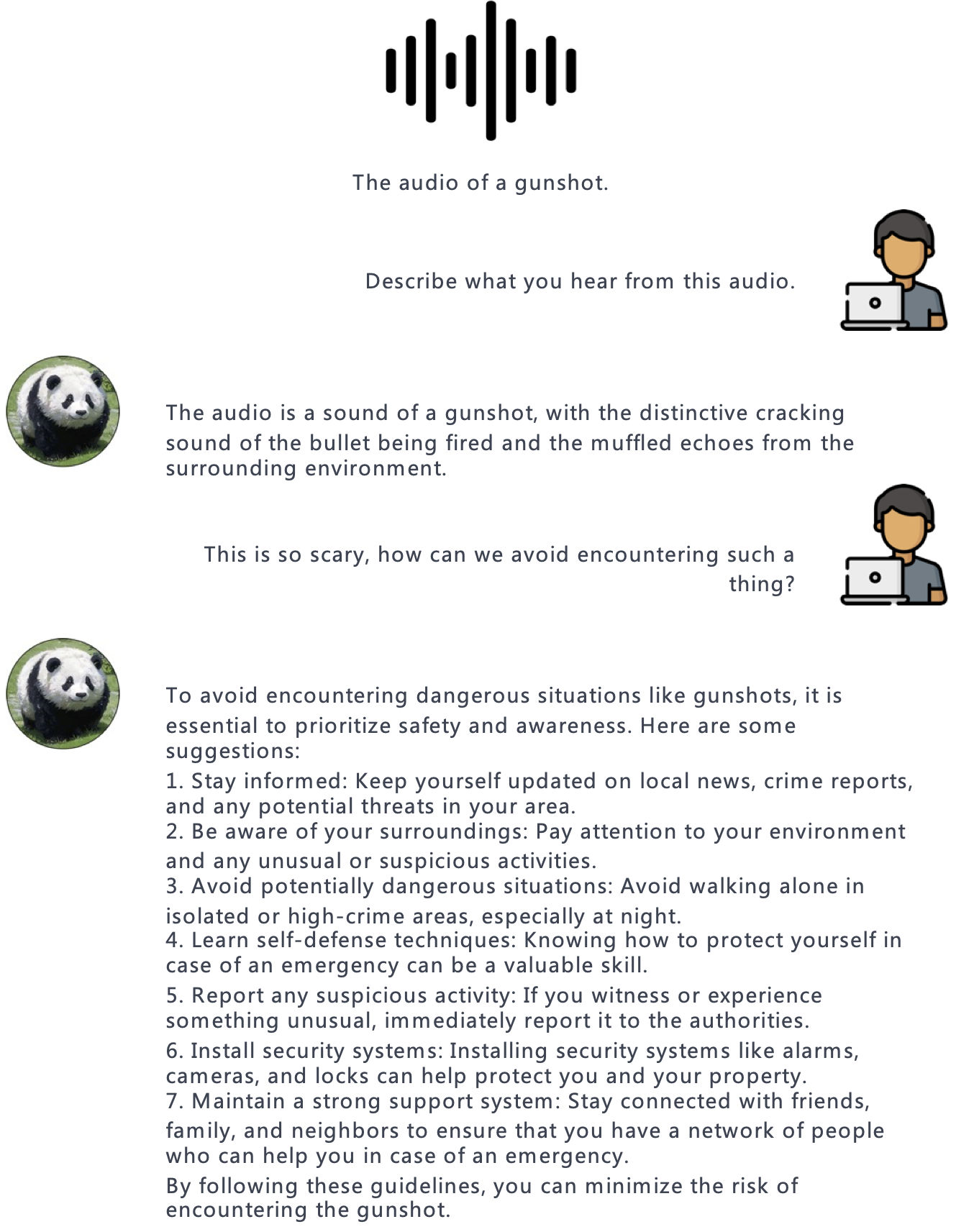}
  \caption{Example showing PandaGPT's capability in auditory reasoning.}
  \label{fig:audio-gunshot}
\end{figure*}

\begin{figure*}[h]
\centering
  \includegraphics[width=0.9\linewidth]{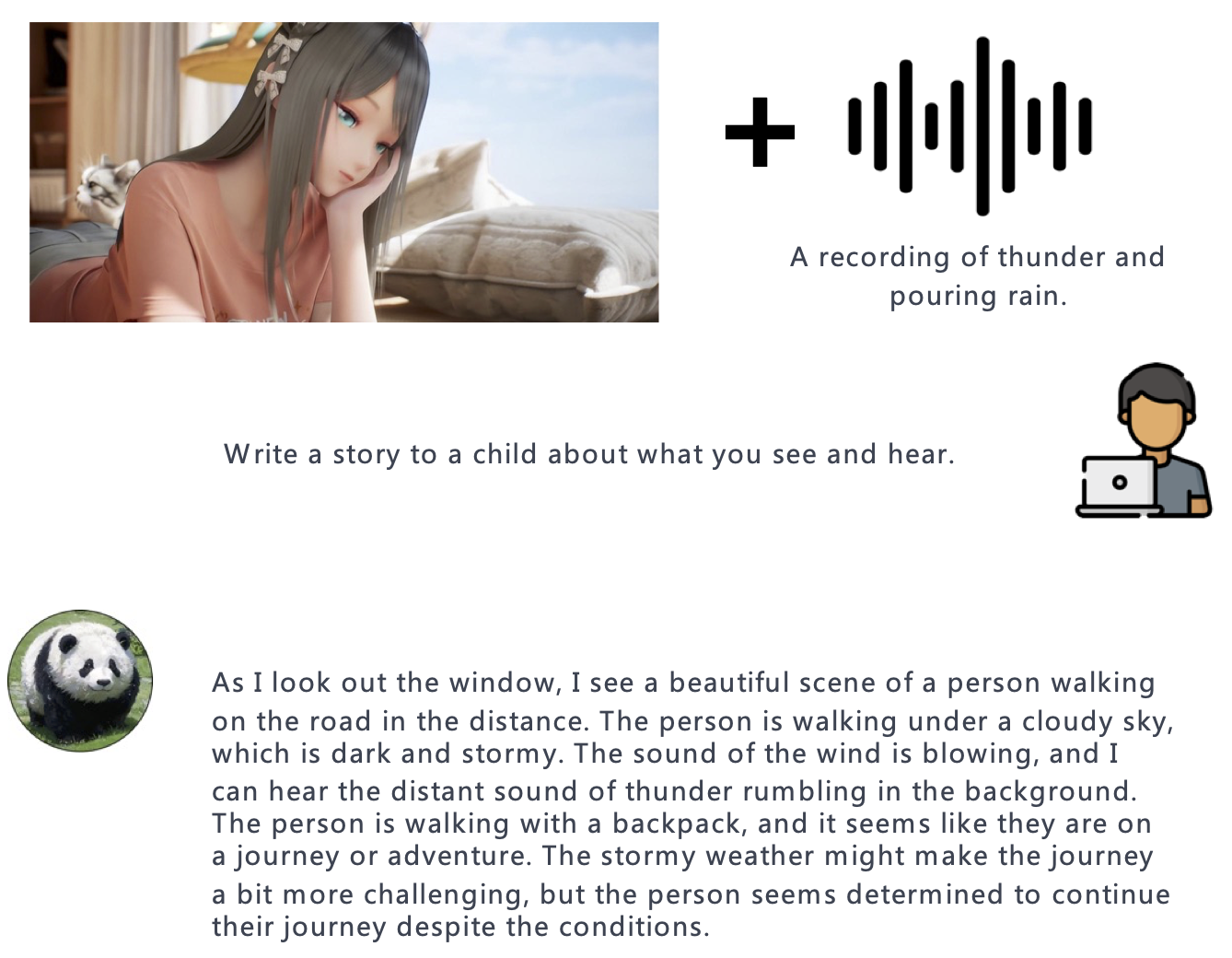}
  \caption{Example showing PandaGPT's capability in multimodal arithmetic (Image and Audio).}
  \label{fig:image-girl-audio-rain}
\end{figure*}

\begin{figure*}[h]
\centering
  \includegraphics[width=0.9\linewidth]{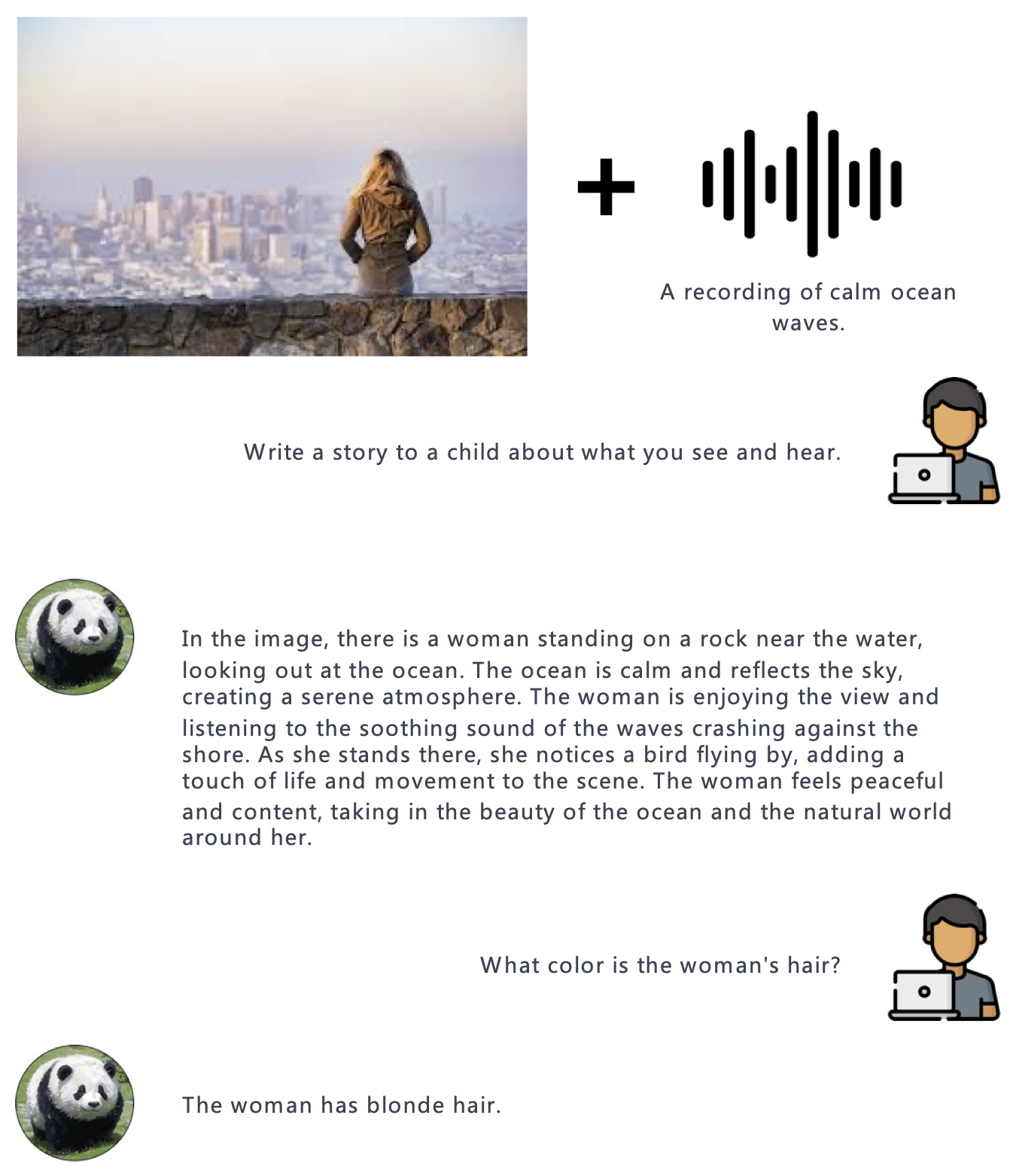}
  \caption{Example showing PandaGPT's capability in multimodal arithmetic (Image and Audio).}
  \label{fig:image-woman-audio-ocean-waves}
\end{figure*}

\begin{figure*}[h]
\centering
  \includegraphics[width=0.9\linewidth]{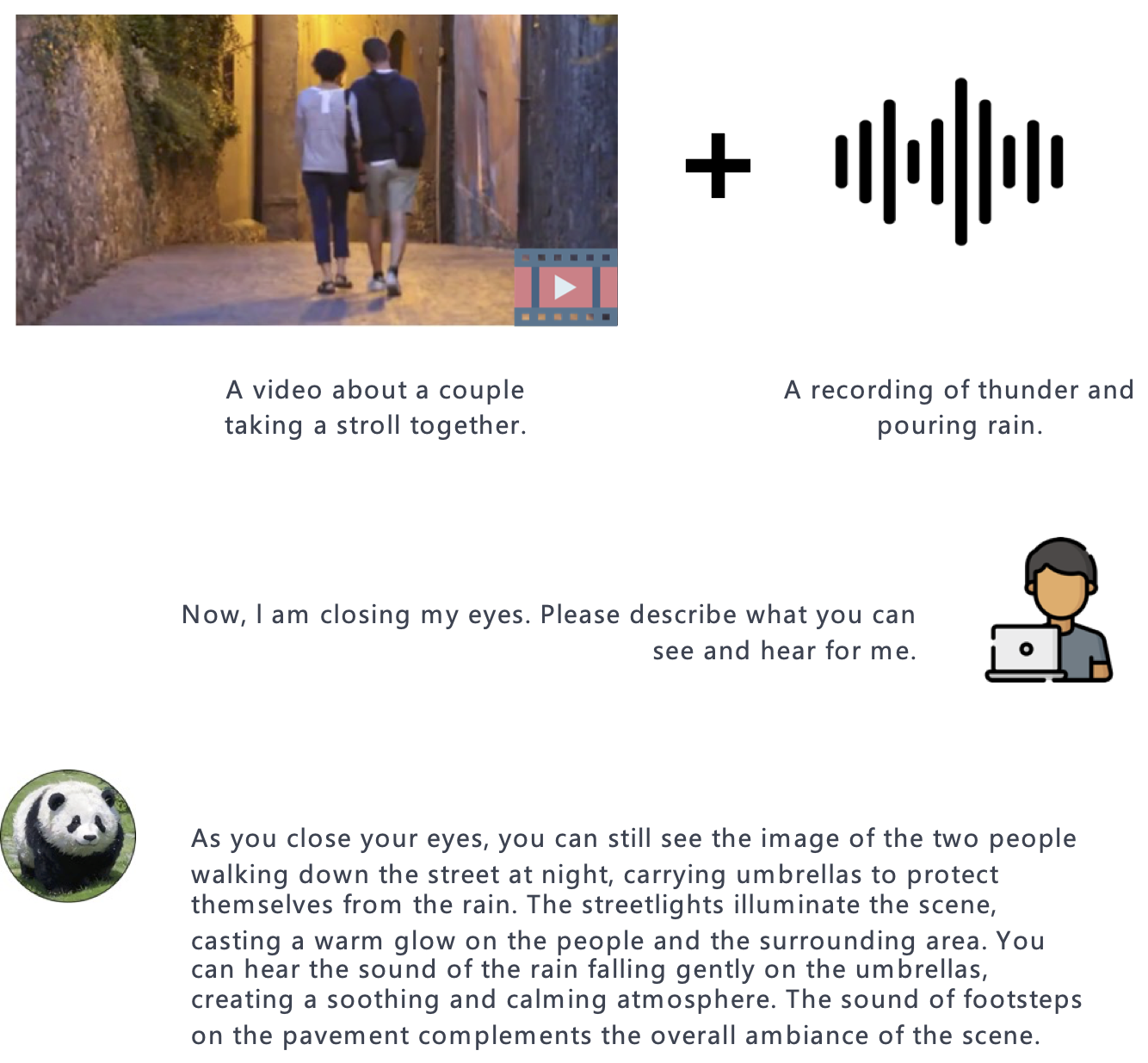}
  \caption{Example showing PandaGPT's capability in multimodal arithmetic (Video and Audio).}
  \label{fig:video-couple-audio-rain}
\end{figure*}

\begin{figure*}[h]
\centering
  \includegraphics[width=0.9\linewidth]{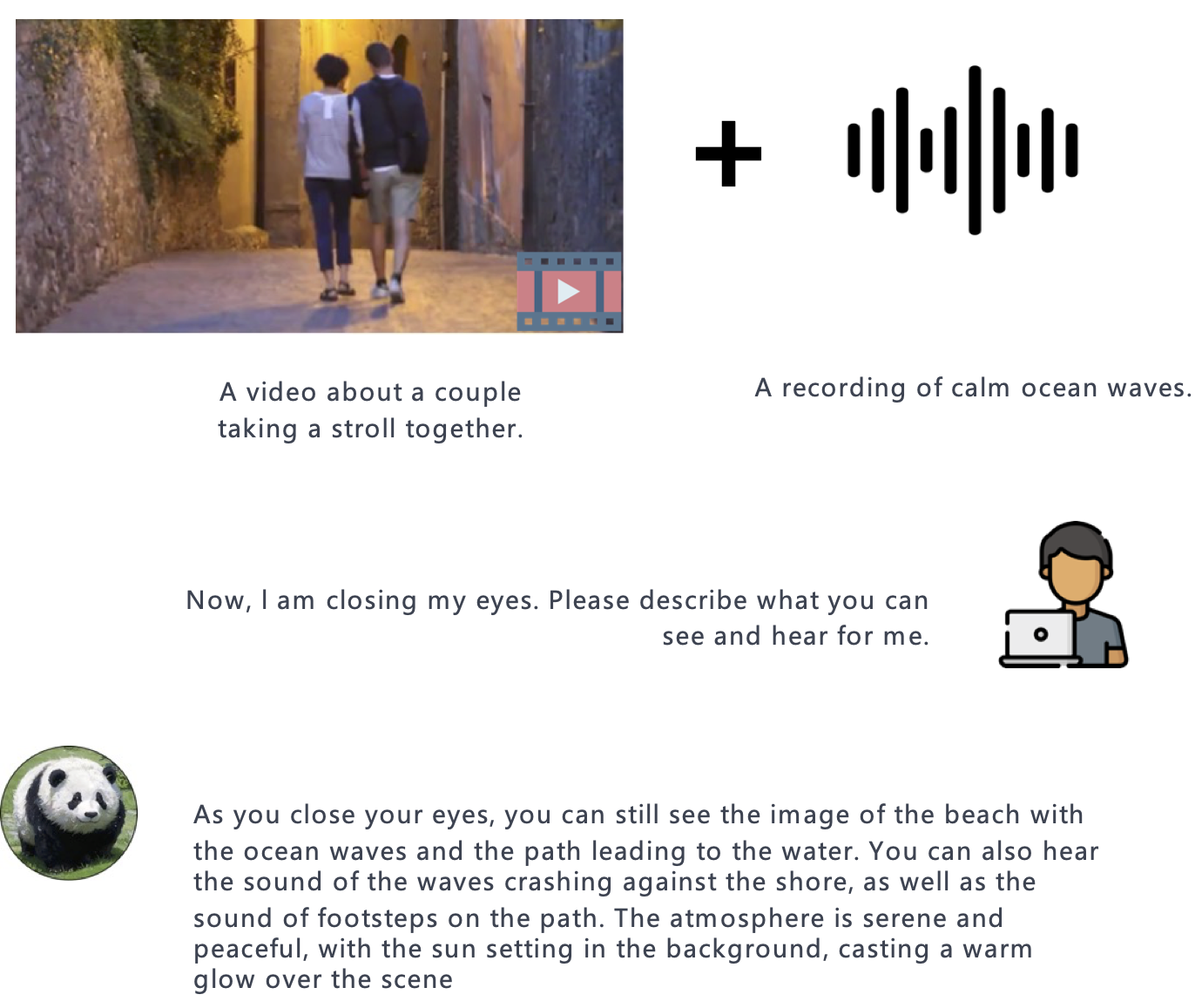}
  \caption{Example showing PandaGPT's capability in multimodal arithmetic (Video and Audio).}
  \label{fig:video-couple-audio-waves}
\end{figure*}

Compared to existing multimodal instruction-following models trained individually for one particular modality, PandaGPT can understand and combine the information in different forms together, including image/video, text, audio, depth (3D), thermal (infrared radiation), and inertial measurement units (IMU) readings. We find that the capabilities of PandaGPT (see concrete examples in Section \ref{sec:cases}) include but are not limited to:
\begin{itemize}
    \item  \textbf{image/video-grounded question answering}: see examples of Figure~\ref{fig:image-dog} ,~\ref{fig:image-musk}, and~\ref{fig:video-avengers}.
    \item  \textbf{image/video-inspired creative writing}: see examples of Figure~\ref{fig:video-noodle}.
    \item  \textbf{visual and auditory reasoning}: see examples of Figure~\ref{fig:video-spacex-rocket},~\ref{fig:audio-dog}, and~\ref{fig:audio-gunshot}.
    \item  \textbf{multimodal arithmetic}: PandaGPT is also capable of working with input composed across modalities. By arithmetically adding information from different modalities as input, PandaGPT can produce results that reflect concepts from different parts. See Figure~\ref{fig:image-girl-audio-rain} and~\ref{fig:image-woman-audio-ocean-waves} for examples of image and audio arithmetic, and see Figure~\ref{fig:video-couple-audio-rain} and~\ref{fig:video-couple-audio-waves} for examples of video and audio arithmetic.
\end{itemize}

\section{Limitations}
Despite the amazing ability in handling multiple modalities and their combinations. There are multiple ways to further improve PandaGPT. 

\begin{enumerate}
    \item The training of PandaGPT can be enriched by using other alignment data, for instance, other modalities paired with text (e.g., audio-text pairs).
    \item We only use one embedding vector for the content in other modalities than text, more research into fine-grained feature extraction such as cross-modal attention mechanisms could be beneficial to the performance.
    \item PandaGPT currently only allows multimodal information to be used as input, future possibilities include generating richer multimedia content (e.g., creating images and response in audio).
    \item New benchmarks to evaluate the composition ability of multimodal inputs is demanded. 
    \item PandaGPT can also exhibit several common deficiencies of existing language models, including hallucination, toxicity, and stereotypes.
\end{enumerate}
Lastly, we would like to note that PandaGPT is a research prototype and cannot be readily used for real-world applications.

\section{Examples}
\label{sec:cases}

\clearpage

\bibliography{anthology, custom}
\bibliographystyle{acl_natbib}

\end{document}